\documentclass[conference]{IEEEtran}
\IEEEoverridecommandlockouts
\usepackage{cite}
\usepackage{amsmath,amssymb,amsfonts}
\usepackage{algorithmic}
\usepackage{graphicx}
\usepackage{textcomp}
\usepackage{xcolor}
\usepackage{hyperref}
\usepackage{bm}
\def\BibTeX{{\rm B\kern-.05em{\sc i\kern-.025em b}\kern-.08em
    T\kern-.1667em\lower.7ex\hbox{E}\kern-.125emX}}
\begin{document}

\title{FAD-SAR: A Novel Fishing Activity Detection System via Synthetic Aperture Radar Images Based on Deep Learning Method}

\author{\IEEEauthorblockN{
Yanbing Bai\IEEEauthorrefmark{1},
Siao Li\IEEEauthorrefmark{1},
Rui-Yang Ju\IEEEauthorrefmark{2},
Zihao Yang\IEEEauthorrefmark{1},
Jinze Yu\IEEEauthorrefmark{3}\IEEEauthorrefmark{5},
Jen-Shiun Chiang\IEEEauthorrefmark{4}
}

\IEEEauthorblockA{\IEEEauthorrefmark{1}
Center for Applied Statistics, School of Statistics, Renmin University of China, Beijing 100872, China}
\IEEEauthorblockA{\IEEEauthorrefmark{2}
Graduate Institute of Networking and Multimedia, National Taiwan University, Taipei City, 106335 Taiwan}
\IEEEauthorblockA{\IEEEauthorrefmark{3}
Department of Mechanical Systems Engineering, Tokyo University of Agriculture and Technology, Tokyo, 183-8538, Japan}
\IEEEauthorblockA{\IEEEauthorrefmark{4}
Department of Electrical and Computer Engineering, Tamkang University, New Taipei City, 251301, Taiwan}
\IEEEauthorblockA{\IEEEauthorrefmark{5}
Corresponding author. E-mail: jinze.yu.kp@gmail.com}}

\maketitle

\begin{abstract}
Illegal, unreported, and unregulated (IUU) fishing activities seriously affect various aspects of human life. However, traditional methods for detecting and monitoring IUU fishing activities at sea have limitations. Although synthetic aperture radar (SAR) can complement existing vessel detection systems, extracting useful information from SAR images using traditional methods remains a challenge, especially in IUU fishing. This paper proposes a deep learning based fishing activity detection system, which is implemented on the xView3 dataset using six classical object detection models: SSD, RetinaNet, FSAF, FCOS, Faster R-CNN, and Cascade R-CNN. In addition, this work employs different enhancement techniques to improve the performance of the Faster R-CNN model. The experimental results demonstrate that training the Faster R-CNN model using the Online Hard Example Mining (OHEM) strategy increases the Avg-F1 value from 0.212 to 0.216.
\end{abstract}

\begin{IEEEkeywords}
remote sensing image, synthetic aperture radar, computer vision, deep learning, object detection, fishing activity detection
\end{IEEEkeywords}

\section{Introduction}
Illegal, unreported, and unregulated (IUU) fishing activities pose a major threat to marine ecosystems, human food supply, and regional political stability.  Specifically, multinational corporations are involved in tax evasion, piracy, and the trafficking of drugs, weapons, and humans, all of which are linked to IUU fishing \cite{belhabib2020illegal}. In addition, IUU fishing exacerbates climate change and severely impacts marine resources \cite{voigt2020oceans}. The research \cite{ndiaye2011illegal} indicates that one-fifth of the world's fishing activities may be illegal or unreported, and the situation is even more serious in less-developed countries or regions. For example, the actual catch is estimated to be 40\% higher than the reported catch in West Africa.

\begin{figure}[t]
\centering
\includegraphics[width=0.95\linewidth]{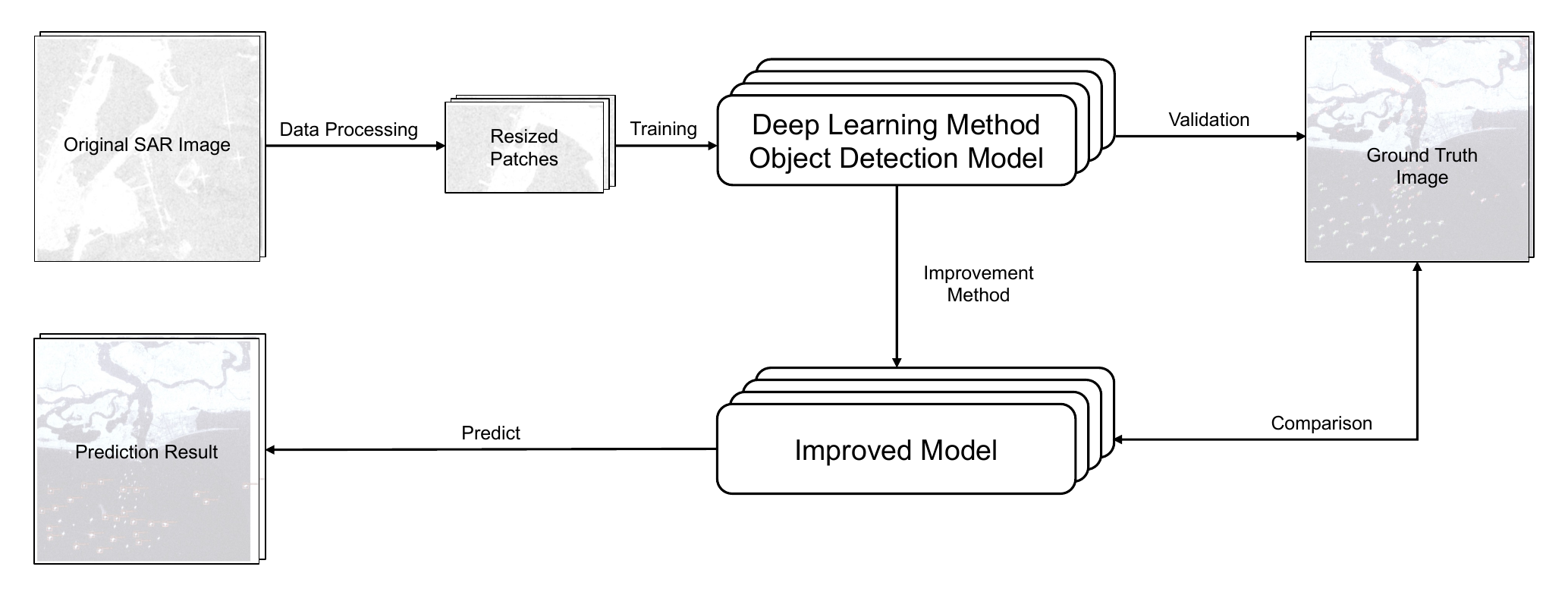}
\caption{The overall architecture of the proposed fishing activity detection system, mainly consists of the training of deep learning models, and the employment of enhancement techniques to improve the model performance.}
\label{figure_system}
\vspace{-4mm}
\end{figure}

With the rapid development of deep learning, neural network models \cite{du2019saliency} are increasingly applied to remote sensing images. Object detection is a hot research topic in the computer vision field, which can be facilitated by classical neural network models \cite{miao2022improved} to help experts analyze synthetic aperture radar (SAR) images, including vessel detection and classification. This paper proposes a novel system for detecting fishing activities using deep learning methods with the aim of detecting potential IUU fishing activities on SAR images, and the overall architecture is shown in Fig. \ref{figure_system}. Specifically, this work conducts experiments on the xView3 dataset \cite{paolo2022xview3}, utilizing six classical object detection models, including Single Shot MultiBox Detector (SSD) \cite{liu2016ssd}, RetinaNet \cite{lin2017focal}, Feature Selective Anchor-Free (FSAF) \cite{zhu2019feature}, Fully Convolutional One-Stage (FCOS) \cite{tian2019fcos}, Faster R-CNN \cite{ren2015faster}, and Cascade R-CNN \cite{cai2018cascade}. In addition, different enhancement techniques \cite{pang2019libra,shrivastava2016training,zhu2019deformable} are used to improve the performance of Faster R-CNN model.

The main contributions of this paper are as follows:
\begin{itemize}    
\item This paper presents a novel system for detecting fishing activities using deep learning methods, which aims to help experts extract information from SAR images more effectively to prevent potential IUU fishing.
\item This work utilizes six classical object detection models on the xView3 dataset for fishing activity detection, which highlights the importance of deep learning methods for detecting IUU fishing.
\item This paper employs the OHEM strategy to train Faster R-CNN with ResNet model to improve the performance.
\end{itemize}

\section{Related Works}
Synthetic aperture radar (SAR) is the high resolution imaging radar that utilizes the principle of synthetic aperture to enhance the angular resolution of the radar, while pulse compression technology improves the range resolution \cite{moreira2013tutorial}. The advancement of SAR technology allows it to provide valuable complementary capabilities to other vessel detection systems \cite{li2021novel}. SAR technology detects vessels at sea based on their strong reflected signals, which are not affected by clouds or weather conditions. In addition, SAR \cite{brown1967synthetic} has been shown as the reliable vessel detection because it is not interfered with by weather and does not rely on vessel cooperation.

However, due to the image resolution, sensor incidence angle, vessel construction materials, radar cross-section, wind or wave conditions, and the lack of ground-truth data, experts face challenges in extracting valuable information from SAR images using traditional methods to validate analytical results. They must classify vessels by type and activity (fishing and non-fishing), assess the possibility of IUU fishing activities, and perform tasks ranging from simple estimates of vessel presence and length to accomplish reliable detection of vessels engaged in IUU fishing.

Therefore, this paper aims to help related experts to extract useful information from SAR images to prevent potential IUU fishing through the proposed deep learning based system for detecting fishing activities.

\section{Method}
\subsection{Dark Vessel Detection System}
The advent of SAR images has introduced a versatile tool capable of detecting vessels that may attempt to evade fisheries enforcement authorities, regardless of weather conditions or time of day. Currently, the European Space Agency's Sentinel-1 satellite has almost global coastline coverage with revisit times of 1 to 12 days. This extensive coverage can be used to recognize vessels engaged in IUU fishing. However, such analysis requires significant automation on a scale sufficient to effectively address IUU fishing. Therefore, this paper proposes a novel fishing activity detection system based on deep learning method, as illustrated in Fig. \ref{figure_system}. The system comprises three main components: data processing, deep learning method, and improvement method, which are detailed in Sections \ref{sec:data processing}, \ref{sec:deep learning}, and \ref{sec:improvement}, respectively.

\subsection{Data Processing}
\label{sec:data processing}
Since the SAR image is large in size, it must be processed into small, three-channel patches before being input to the neural network.
The SAR images are cut into $800 \times 800$ pixel patches and the size of the patches follows the original settings \cite{paolo2022xview3}.
In terms of channels, vertical-vertical (VV) and vertical-horizontal (VH) each has its own advantages: VH is generally more suitable for detecting vessels as it offers a better contrast between vessel and ocean clutter, while VV provides detailed information about sea surface elements.
Utilizing VV and VH as two channels, the third channel offers multiple options, including the use of five auxiliary images or the duplication of the two SAR channels.  
However, due to the use of min-max normalization, channels with the same slice value are considered as NaN and must be removed.

This paper introduces several fusion methods: single auxiliary channel, mean embedding of VV and VH channels, difference embedding of VV and VH channels, mean embedding of auxiliary channels, and mean embedding of all channels. 
However, experiments indicate that using auxiliary channels does not produce any improvement. 
Channel visualization shows that the spatial resolution of the auxiliary data is significantly lower than that of the SAR data and provides poorer information than the SAR images. 
Channel concatenation requires upsampling of low resolution data, which leads to biased object locations and may mislead training.
Other possible reasons for the lack of improvement include the relatively low importance of the bathymetric signals, the ability of the model to implicitly infer wind speed and direction from the SAR images, and model performance limitations due to the quality of the labels rather than the lack of information in the input data.
Therefore, this paper chooses to use the average embedding of the VV and VH channels as the input for the third channel. Although this does not provide additional information, it balances the advantages of the two channels and thus enhances system stability.

\subsection{Deep Learning Method}
\label{sec:deep learning}
This work trains different neural networks for fishing activity detection including SSD \cite{liu2016ssd}, RetinaNet \cite{lin2017focal}, FSAF \cite{zhu2019feature}, FCOS \cite{tian2019fcos}, Faster R-CNN \cite{ren2015faster}, and Cascade R-CNN \cite{cai2018cascade}.

One-stage object detection models include SSD \cite{liu2016ssd}, RetinaNet \cite{lin2017focal}, FSAF \cite{zhu2019feature}, and FCOS \cite{tian2019fcos}. 
Specifically, SSD \cite{liu2016ssd} uses multiple anchors on feature maps at different scales to detect objects. SSD predicts bounding boxes and classes directly in a single network, balancing speed and accuracy. 
RetinaNet \cite{lin2017focal} employs Feature Pyramid Network (FPN) \cite{lin2017feature} and Focal Loss to solve the class imbalance problem. Focal Loss reduces the penalty for simple negative samples and allows the model to focus more on difficult to classify samples, thus enhancing the model performance. 
Feature Selective Anchor-Free (FSAF) \cite{zhu2019feature} dynamically selects the most suitable feature layer for each object from the multi-scale features for prediction. This method eliminates the complex anchor matching process in traditional anchor-based methods, simplifying the training process and improving performance. 
Fully Convolutional One-Stage (FCOS) \cite{tian2019fcos} eliminates the anchor point box design and directly predicts the four offsets and class probabilities of the bounding box at each location. This design simplifies the model structure and enables flexible adaptation to objects of varying shapes and sizes.

Two-stage object detection models include Faster R-CNN \cite{ren2015faster} and Cascade R-CNN \cite{cai2018cascade}.
Faster R-CNN \cite{ren2015faster} is the modified version of Fast R-CNN \cite{girshick2015fast}, which integrates the region proposal network (RPN) with the convolutional neural network (CNN) to generate region proposals at little additional cost. The RPN shares complete image convolutional features with the detection network, greatly reducing the cost of generating region suggestions, while the full convolutional network (FCN) \cite{long2015fully} predicts the boundaries of two objects at each location.
Cascade R-CNN \cite{cai2018cascade} extends R-CNN \cite{girshick2014rich} by incorporating multiple cascaded detector stages that become increasingly selective for tight false positive examples. Each stage adjusts the bounding box to find a suitable set of tight false positive examples to train the next stage, effectively addressing the overfitting problem and enabling efficient training. This cascading process also applies to the inference process, consistent with the assumption of incremental improvement through higher quality detectors at each stage.

\subsection{Improvement Method}
\label{sec:improvement}
This paper employs enhancement techniques to improve the performance, including Data Augmentation (DA), Deformable ConvNets v2 (DCNv2) \cite{zhu2019deformable}, IoU-Balanced Sampling (IoU-BS) \cite{pang2019libra}, and Online Hard Example Mining (OHEM) \cite{shrivastava2016training}.

DCNv2 \cite{zhu2019deformable} innovatively introduces deformable convolutional operations. Traditional convolutional operations are bound by fixed kernels when extracting features, which limit their effectiveness in dealing with complex scenarios such as object deformation and occlusion.
Deformable convolution allows the convolution kernel to dynamically adjust its shape and position on the feature map, combining kernel weights and offset values to better align with the object's features.
This flexibility enhances DCNv2's ability to represent complex scenes involving object deformation and occlusion. By integrating deformable convolutional modules into the network, the accuracy and robustness of object detection can be significantly improved. This adaptability enables DCNv2 to effectively capture object deformation information, which enhances the overall robustness and accuracy of the model.

IoU-BS \cite{pang2019libra} takes samples based on IoU values. Traditional sampling methods often use random sampling or a fixed ratio of positive and negative samples, which may cause the model to over-prioritize easy samples and ignore hard-to-detect objects. 
In addition, IoU-BS is able to assign weights to positive and negative samples based on IoU values, which makes the model focus more on challenging objects. This method achieves a better balance in the training set by adjusting the proportion of positive and negative samples, which can enhance the performance and robustness of the model. This sampling strategy effectively improves the model's ability to recognize objects in complex scenes. 

OHEM \cite{shrivastava2016training} can dynamically select challenging samples to emphasize during training. Its innovation lies in the use of online hard-case mining strategy. Unlike traditional training methods that treat all samples equally, for OHEM, some samples may be more difficult to detect in real-world scenarios due to variations in object size, pose, and other factors. OHEM enhances the model's ability to recognize these challenging objects by filtering out samples with higher loss function values in each round of training and prioritizing the training of these difficult samples. This method can significantly enhance the performance and generalization ability of the model, enabling it to better cope with challenges in complex scenarios.

\section{Experiments}
\label{sec:experiments}
\subsection{Dataset}
The xView3 dataset \cite{paolo2022xview3} comprises SAR images and data for marine object recognition. It consists of approximately 1,000 scenes from the ocean region, each scene consisting of seven images: two SAR images with different polarization signatures (VV, VH) and five auxiliary images (bathymetry, wind speed, wind direction, wind mass, and land/ice cover).

The prediction tasks applied to this dataset are divided into three categories: recognizing marine objects in each scene, estimating the length of each object and judging if it is a vessel, and classifying each vessel as either a fishing vessel or not. In addition, data annotations and labels integrate vessel detection from the Global Fishing Watch (GFW) model and AIS/VMS records to provide high-quality labeled data specific to marine object detection. Specific data labels include the object's latitude and longitude, vessel length, source of record, vessel classification (vessel or non-vessel, fishing vessel or non-fishing vessel), distance from the coastline, and label confidence level.

\begin{table*}[t]
\centering
\caption{Quantitative comparison with different classic object detection models for fishing activity detection on the xView3 dataset}
\label{tab:comparison}
\setlength{\tabcolsep}{6.4mm}{
\begin{tabular}{llccccc}
\hline \noalign{\smallskip}
\textbf{Detector} & \textbf{Backbone} & \bm{$F1_D$} & \bm{$F1_S$} & \bm{$F1_V$} & \bm{$F1_F$} & \bm{$Avg-F1$} \\ \noalign{\smallskip} \hline \noalign{\smallskip}
SSD \cite{liu2016ssd} & VGG16 \cite{simonyan2014very} & {\color{blue}0.49551} & {\color{red}0.13811} & 0.79081 & 0.49043 & 0.20295 \\
FCOS \cite{tian2019fcos} & ResNet50 \cite{he2016deep} & 0.33735 & 0.10571 & 0.46333 & 0.10648 & 0.06312 \\
FSAF \cite{zhu2019feature} & ResNet50 \cite{he2016deep}& 0.41608 & 0.08624 & 0.66950 & 0.29027 & 0.12053 \\
RetinaNet \cite{lin2017focal} & ResNet50 \cite{he2016deep} & 0.44654 & 0.07494 & 0.77979 & 0.41053 & 0.15518 \\
Cascade R-CNN \cite{cai2018cascade} & ResNet50 \cite{he2016deep} & 0.49278 & {\color{blue}0.13180} & {\color{blue}0.79479} & {\color{blue}0.51454} & {\color{blue}0.20445} \\
Faster R-CNN \cite{ren2015faster} & ResNet50 \cite{he2016deep} & {\color{red}0.51246} & 0.10335 & {\color{red}0.82152} & {\color{red}0.55647} & {\color{red}0.21215} \\ \noalign{\smallskip} \hline \noalign{\smallskip}
\multicolumn{7}{l}{Best and 2nd best performance are in {\color{red}red} and {\color{blue}blue} colors, respectively.}
\end{tabular}}
\vspace{-4mm}
\end{table*}

\begin{table}[ht]
\centering
\caption{Ablation study of the effect of bounding size on performance}
\label{tab:ablation}
\begin{tabular}{cccccc}
\hline \noalign{\smallskip}
\begin{tabular}[c]{@{}c@{}}\textbf{Bounding}\\\textbf{Size}\end{tabular} & \bm{$F1_D$} & \bm{$F1_S$} & \bm{$F1_V$} & \bm{$F1_F$} & \bm{$Avg-F1$} \\ \noalign{\smallskip} \hline \noalign{\smallskip}
$10\times10$ & 0.01154 & 0.00496 & {\color{red}0.88235} & {\color{red}0.66667} & 0.00639 \\
$20\times20$ & {\color{red}0.51246} & {\color{blue}0.10335} & {\color{blue}0.82152} & {\color{blue}0.55647} & {\color{red}0.21215} \\
$30\times30$ & 0.50531 & 0.09895 & 0.79979 & 0.50163 & 0.19660 \\
$40\times40$ & {\color{blue}0.50591} & {\color{red}0.10580} & 0.81532 & 0.50458 & {\color{blue}0.20185} \\ \noalign{\smallskip} \hline \noalign{\smallskip}
\multicolumn{6}{l}{Best and 2nd best performance are in {\color{red}red} and {\color{blue}blue} colors, respectively.}
\end{tabular}
\vspace{-4mm}
\end{table}

\subsection{Evaluation Metric}
\subsubsection{$F-score$}
F-score is a widely used metric for evaluating the precision of algorithms, as it considers both precision and recall, providing a balanced reflection of the performance of the algorithm. The equation for the F-score is as:
\begin{equation}
\label{f-score}
F\!-\!score = \frac{\left(1+\beta^{2}\right) \times Precision \times Recall}{\beta^{2} \times Precision + Recall}
\end{equation}
where $\beta$ is the parameter that determines the precision weight in the harmonic mean. In this work, we specifically use the F1-score, where $\beta=1$, representing the harmonic mean of precision and recall.

\subsubsection{$F1-Score$}
Typically, precision and recall are equally weighted in the F1-score, representing the harmonic mean of precision and recall. The formula is as follows:
\begin{equation}
\begin{aligned}
\label{f1-score}
F1\!-\!score &= \frac{2 \times Precision \times Recall}{Precision + Recall} \\
&= \frac{2T_P}{2T_P+F_P+F_N}
\end{aligned}
\end{equation}
where $T_P$, $F_P$, and $F_N$ represent true positives, false positives, and false negatives, respectively.

\subsubsection{$F1_D$, $F1_S$, $F1_V$, and $F1_F$}
This work calculates each of the following four aggregated metrics:
\begin{itemize}
\item $F1_D$: F1-score for maritime object detection
\item $F1_S$: F1-score for close-to-shore object detection
\item $F1_V$: F1-score for vessel classification
\item $F1_F$: F1-score for fishing classification
\end{itemize}

\subsubsection{$Avg-F1$}
In the previous challenge \cite{paolo2022xview3}, the Overall Ranking Metrics used by the organizers are calculated from $F1_D$, $F1_S$, $F1_V$, $F1_F$, and $PE_L$ (aggregate percent error about vessel length estimation). Therefore, this work cannot directly use the Overall Ranking Metrics from the challenge \cite{paolo2022xview3} to evaluate the model performance. We propose a metric that is more suitable for this experiment to evaluate the model performance:
\begin{equation}
\label{avg-f1}
Avg-F1=\frac{F1_D+F1_S}{2}\times\frac{F1_V+F1_F}{2}
\end{equation}

\subsection{Experiment Setup}
All experiments for this work are conducted on the machine learning cloud platform with Intel Core i7-9700K CPUs and NVIDIA GeForce RTX 2080 Ti GPUs (11GB memory for each GPU) supported by 32 GB RAM.

\subsection{Training}
To fairly compare the performance of different models, all models in this work use the same parameter settings.
We utilize RetinaNet \cite{lin2017focal}, FSAF \cite{zhu2019feature}, FCOS \cite{tian2019fcos}, Faster R-CNN \cite{ren2015faster}, and Cascade R-CNN \cite{cai2018cascade} architectures with ResNet50 \cite{he2016deep} as the backbone. However, for SSD \cite{liu2016ssd}, we employ the default settings of VGG16 \cite{simonyan2014very}.
Data loading is accomplished using the same pipeline, except for the data augmentation method. The batch size is set to 8, and we utilize the SGD optimizer. The initial learning rate is set to 0.02; the momentum is 0.9, and the weight decay is set to 0.0001.
For the learning rate tuning strategy, we use the warm-up ratio of 0.0001 for the first 1000 iterations, followed by a multistep learning rate scheduling strategy with a step size of 3 and a gamma value of 0.3.
Due to hardware constraints, some parameters (e.g., the batch size of the Cascade R-CNN) are adapted specifically to fit the function of available computing resources. Specifically, we adjust the batch size of Cascade R-CNN to 4 to prevent GPU memory overflow, and the initial learning rate of SSD and FCOS to 0.002 to mitigate the gradient explosion problem.

\subsection{Ablation Study}
In terms of labeling, both class labels and bounding box labels required reprocessing. 
The class labels ``is\_vessel'' and ``is\_fishing'' are merged and converted into three types of labels: ``fishing'', ``non\_fishing'', and ``non\_vessel''. 
During data processing, we encounter a significant number of missing labels that could not be classified correctly, primarily from low to medium confidence labels.
Given the complexity of pseudo-labelling and based on Pillai \emph{et al.}'s research \cite{pillai2022deepsar} on how to deal with the label noise, this paper only selects the filtered high-confidence labels to address the label noise problem.
The original labels only contain sample centers of mass, so that adjusting the bounding box size is necessary if bounding box labels are required.
Therefore, we conduct an ablation study using ResNet50 \cite{he2016deep} as the backbone and Faster R-CNN \cite{ren2015faster} as the architecture, and the results are presented in Table \ref{tab:ablation}, indicating that the optimal performance is attained with a bounding box size of $20 \times 20$, which is consequently selected as the final size.

\begin{table}[t]
\centering
\caption{Quantitative comparison of different enhancement techniques of Faster R-CNN with ResNet 50 model on the xView3 dataset}
\label{tab:method}
\setlength{\tabcolsep}{1.5mm}{
\begin{tabular}{lccccc}
\hline \noalign{\smallskip}
\textbf{Method} & \bm{$F1_D$} & \bm{$F1_S$} & \bm{$F1_V$} & \bm{$F1_F$} & \bm{$Avg-F1$} \\ \noalign{\smallskip} \hline \noalign{\smallskip}
Baseline & {\color{blue}0.51246} & 0.10335 & {\color{blue}0.82152} & {\color{blue}0.55647} & {\color{blue}0.21215} \\
DA & 0.50099 & 0.11729 & 0.79087 & 0.52184 & 0.20291 \\
DCNv2 \cite{zhu2019deformable} & 0.50315 & {\color{blue}0.12667} & 0.80083 & 0.51775 & 0.20762 \\
IoU-BS \cite{pang2019libra} & 0.49637 & {\color{red}0.13549} & 0.77406 & 0.53786 & 0.20724 \\
OHEM \cite{shrivastava2016training} & {\color{red}0.51167} & 0.10514 & {\color{red}0.82754} & {\color{red}0.57521} & {\color{red}0.21631} \\ \noalign{\smallskip} \hline \noalign{\smallskip}
\multicolumn{6}{l}{Best and 2nd best performance are in {\color{red}red} and {\color{blue}blue} colors, respectively.}
\end{tabular}}
\vspace{-4mm}
\end{table}

\subsection{Experimental Results}
Due to the limited prior research on deep learning methods utilizing the xView3 dataset \cite{paolo2022xview3}, we propose a novel fishing activity detection system. This system employs different deep learning models (including one-stage and two-stage object detection models \cite{cai2018cascade,lin2017focal,liu2016ssd,ren2015faster,tian2019fcos,zhu2019feature}) to investigate their model performances on the xView3 dataset.

The quantitative evaluation results of various object detection models on the xView3 dataset are summarized in Table \ref{tab:comparison}.
Among them, the Avg-F1 value of Cascade R-CNN \cite{cai2018cascade} and Faster R-CNN \cite{ren2015faster} are 0.204 and 0.212 respectively, which outperform the one-stage object detection models SSD \cite{liu2016ssd} and RetinaNet \cite{lin2017focal} of 0.203 and 0.155 respectively.
This indicates that the two-stage object detection models usually outperform the one-stage object detection models on the xView3 dataset.
Furthermore, the performance of Cascade R-CNN \cite{cai2018cascade} is poorer compared to Faster R-CNN \cite{ren2015faster}, so that in the following experiments, we perform the enhancement techniques on the Faster R-CNN \cite{ren2015faster} model.
In addition, both FSAF \cite{zhu2019feature} and FCOS \cite{tian2019fcos} models using the anchor-free algorithm perform poorly, indicating that the anchor-free algorithm is not suitable for the xView3 dataset.

\begin{figure}[t]
\centering
\includegraphics[width=\linewidth]{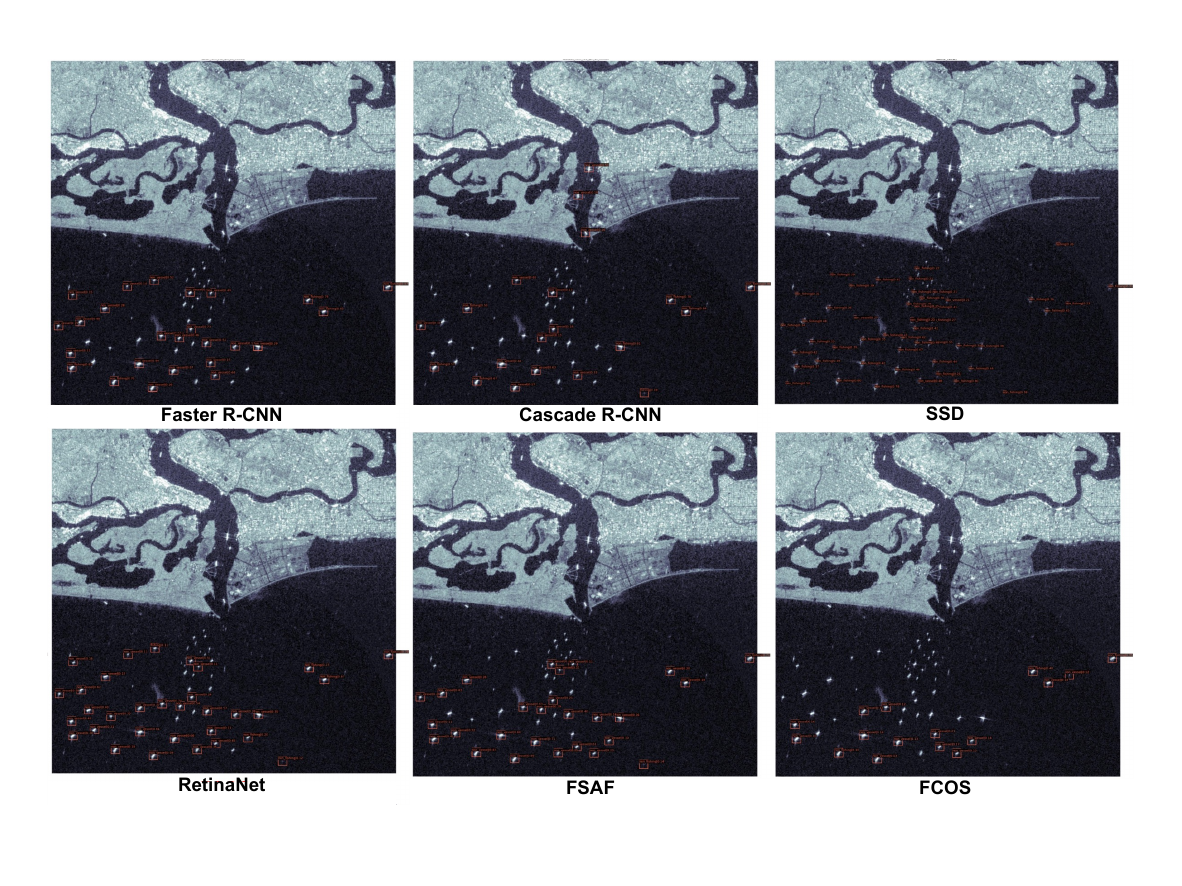}
\caption{Example results of different models for fishing activity detection.}
\label{figure_result}
\vspace{-4mm}
\end{figure}

In the initial stage of data processing, this work employs a simple filtering method that excludes high-confidence labels by retaining only those labels. Therefore, the capabilities of the models to generalize to objects near the coastline are impaired. As shown in Fig. \ref{figure_result}, this limitation is evident in the significant reduction in the $F1_S$ scores assigned to objects near the coast, and therefore also affects the $F1_D$ scores.

Table \ref{tab:method} presents quantitative evaluation results of various improvement methods applied to train the Faster R-CNN \cite{ren2015faster} model, demonstrating that using the OHEM strategy improves the model performance. The Avg-F1 value of the proposed model increases from 0.212 (baseline model) to 0.216, which presents an increase of 1.96\%. It is worth noting that the proposed model achieves the best performance compared to other models, with an Avg F1 value of 0.216.

\section{Conclusion}
It is widely known that extracting useful information from SAR images faces challenges, including image resolution limitations, variations in sensor incidence angles, differences in vessel construction materials, variations in radar cross-section, and wind/wave conditions.
In addition, the lack of ground-truth data poses challenges in corroborating analysis results.
The excellent performance of different classical object detection models on the xView3 dataset highlights the importance of deep learning mehotd in solving the problem of IUU fishing.
Therefore, in this paper, we propose a novel fishing activity detection system that aims to help experts to extract information efficiently. Specifically, the proposed model employs OHEM strategy to train Faster R-CNN models for fishing activity detection.

\bibliographystyle{IEEEtran}
\bibliography{mybibliography}
\end{document}